%% file: symposium2018_optim.tex
\documentclass[runningheads,a4paper]{llncs}

\usepackage{amsmath}
\usepackage{graphicx}            
\usepackage[caption=false]{subfig}
\usepackage{amssymb}             
\usepackage{amsfonts}            
\usepackage{enumitem}            
\usepackage{multirow}            
\usepackage{siunitx}             
\usepackage{url}                 
\usepackage{xspace}              
\usepackage[T1]{fontenc}         
\usepackage{hyperref}            
\usepackage{wrapfig}
\usepackage{todonotes}
\usepackage{units}
\usepackage{subfig}
\usepackage{lipsum}
\usepackage[bottom]{footmisc}    
\usepackage{booktabs,dcolumn}    
\usepackage[firstpage=true]{background}

\newcolumntype{d}[1]{D{.}{.}{#1}} 

\usepackage[%
square,        
comma,         
numbers,       
sort&compress, 
]{natbib}

\graphicspath{{images/}}
\DeclareGraphicsExtensions{.pdf,.png,.jpg,.jpeg}

\DeclareMathOperator*{\argmax}{arg\,max}  

\newcommand{\figlabel}[1]{\label{fig:#1}}

\newcommand{\cm}{CM730\xspace}

\newcommand{\iguhop}{igus\textsuperscript{\tiny\circledR}$\!$ Humanoid Open Platform\xspace}

\newcommand{\eg}{e.g.,\ }

\newcommand{\ie}{i.e.,\ }

\newcommand\copyrighttext{%
	\parbox{\textwidth}{
		\footnotesize
		\textbf{Accepted final version.} In \textit{Proceedings of 22th RoboCup International Symposium, Montreal, Canada.}
	}
}

\SetBgContents{\copyrighttext}
\SetBgScale{1}
\SetBgColor{black}
\SetBgAngle{0}
\SetBgOpacity{1}
\SetBgPosition{current page.south}
\SetBgVshift{4cm}
\SetBgHshift{3mm}

\begin{document}

\mainmatter

\title{Combining Simulations and Real-robot Experiments for Bayesian Optimization of Bipedal Gait Stabilization}
\titlerunning{Bayesian Optimization of Bipedal Gait Stabilization}

\author{Diego Rodriguez\thanks{Both authors contributed equally.}, Andr\'{e} Brandenburger*, and Sven Behnke}
\authorrunning{Rodriguez, Brandenburger, and Behnke}

\institute{Autonomous Intelligent Systems, Computer Science, Univ.\ of Bonn, Germany\\
\url{{rodriguez, behnke}@ais.uni-bonn.de}, \url{andre.brandenburger@uni-bonn.de}
\url{http://ais.uni-bonn.de}}

\maketitle
\begin{abstract}
Walking controllers often require parametrization which must be tuned according to some cost function.
To estimate these parameters, simulations can be performed which are cheap but do not fully represent reality.
Real-robot experiments, on the other hand, are more expensive and lead to hardware wear-off.
In this paper, we propose an approach for combining simulations and real experiments to learn gait stabilization parameters.
We use a Bayesian optimization method which selects the most informative points in parameter space to evaluate based on the entropy of the cost function to optimize.
Experiments with the \iguhop demonstrate the effectiveness of our approach.
\end{abstract}

\input{introduction.tex}

\input{related_work.tex}
\input{background.tex}
\input{approach.tex}
\input{experiments.tex}

\input{conclusions.tex}

\subsection*{Acknowledgements}\vspace*{-1ex}
\footnotesize
The authors would like to thank Alonso Marco from the Max Planck Institute for Intelligent Systems for providing the MF-ES implementation.
This work was partially funded by grant BE 2556/13 of German Research Foundation.


\renewcommand{\bibsection}{\section*{References}}

\begingroup
\small 
\bibliography{symposium2018_optim.bib}
\endgroup

\end{document}

%% file: introduction.tex
\section{Introduction}
\begin{figure}[!b]
	\centering
	\includegraphics[width=\linewidth]{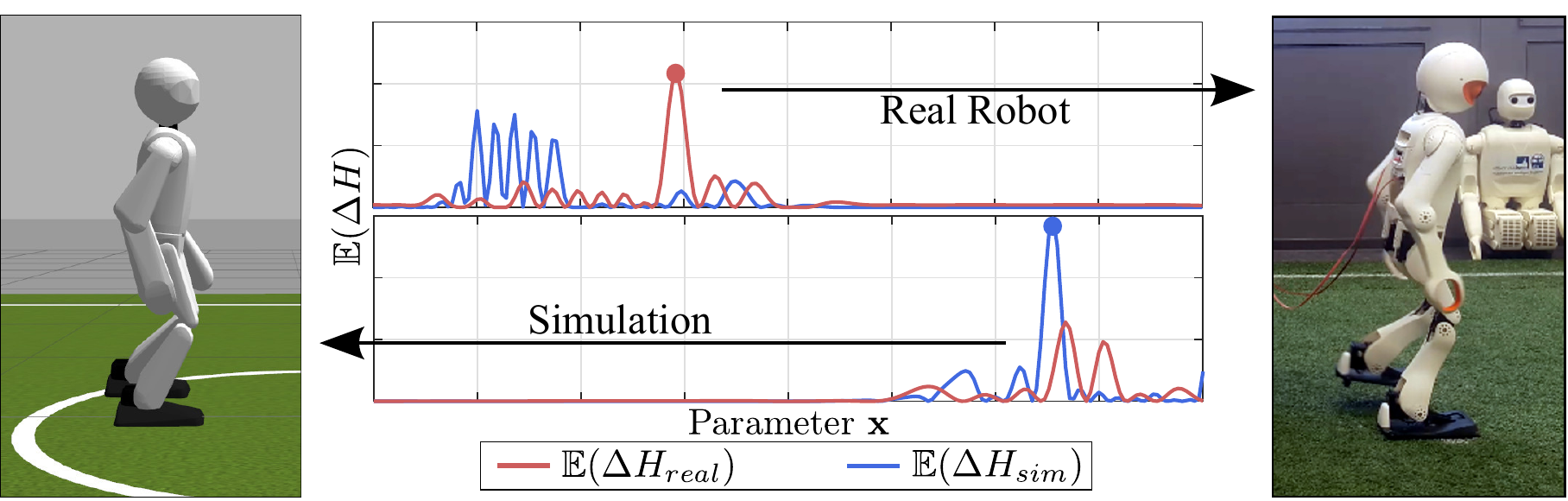}
	\caption{Combining multiple sources of information for learning gait parameters $\mathbf{x}$. 
		Based on the largest relative entropy $\mathbb{E}(\Delta H)$ of a cost function $J$, 
		a walking sequence is performed in simulation or with the real robot.
		The number of walking sequences performed with the real robot is reduced by using simulations.
		}
	\figlabel{introduction}
\end{figure}

Walking is a crucial task for legged robots. 
The state-of-the-art walking controllers and generators typically require a fine-tuned parametrization that due to its complexity is determined by experts.
This puts a constraint on the applicability of these methods.
They can be used mainly by the people who designed them.
In recent years, learning approaches have been proposed in order to reduce the amount of work and expert knowledge required to tune these methods \cite{calandra, atrias}.
This implies thus to perform experiments to estimate parameters.
Consequently, sample-efficient learning approaches are required in order to reduce the hardware wear-off induced by the experiments.
One way to reduce the need for real-robot experiments is to use simulations, which are cheaper to perform but do not fully represent reality.
This point is particularly relevant for low-cost robots whose hardware is not as precise as expected.
In this paper, we propose a method to combine simulations and real-robot experiments to optimize gait parameters, 
specifically to learn activation values of corrective actions that act on top of an open-loop bipedal gait generator. 
The optimization uses a state-of-the-art sample-efficient Bayesian method which selects the most informative points in parameter space to evaluate based on the entropy of a cost function.

%% file: related_work.tex
\section{Related Work}
Several learning methods have been used to optimize manipulation or locomotion parameters \cite{roefer, calandra, deisenroth, atrias, localbayes, heijmink}.
Most of the methods are based on Bayesian optimization---due to its high sample efficiency.
\citet{deisenroth}, for example, developed a Bayesian approach to tune a cart-pole system,
whereas \citet{berkenkamp} proposed to use Bayesian optimization to safely tune robotic controllers for quadrotors.
Moreover, \citet{lqrtune} combined Bayesian optimization with optimal control to tune LQR regulators. 
In contrast to these approaches, \citet{localbayes} suggested to direct the optimization process by using a search distribution,
however, the optimization loses expressibility on a global scope since it only optimizes locally.
Even though these methods take advantage of the sample efficiency of the Bayesian optimization, the robotic hardware is worn off unnecessarily in the learning process, especially in the initial stages where the controllers do not possess any prior knowledge. 

Specifically for the problem of gait parameter optimization, \citet{calandra} suggested
a Bayesian optimization in order to replicate a given target trajectory for a bipedal robot. 
Based on only real experiments, the algorithm was able to find a stable gait. 
Because only the real hardware was used for all experiments, we expect a considerable wear-off of the robot.
On the other hand, \citet{heijmink} proposed a method to learn gait parameters and impedance profiles in simulation for a quadruped robot.
This was accomplished by using the PI\textsuperscript{2} algorithm with a cost function consisting of speed tracking, energy consumption, joint limits, and torques. 
Although the results were validated in the real hardware, no transfer between simulation and the real hardware was addressed.
Similarly, \citet{hengst}, use a simulator to learn gait parameters that will be tested in the real robot. 
This is a reinforcement learning approach 
that learns the ankle joint position of the stance leg and the placement of the swing foot.

There exist several approaches that have transferred knowledge gathered in simulation to real robotic platforms.
\citet{farchy}, for example, learn several dynamic parameters of a simulator in order to get a similar performance compared with real-robot experiments. 
This approach was extended by \citet{hanna} by learning the dynamics of the simulator using the differences of the actions between the real world and the simulation.
The walk velocity of the NAO robot was increased by 43\% starting from a state-of-the-art walk engine.
Nevertheless, 
a human expert was required to select the appropriate parameters to be learned.
Additionally, \citet{cutler} uses a simulator to learn a nonparametric prior that will parametrize a learning algorithm that acts directly on the real platform, 
in other words, there is an one-step transfer from simulation to the real robot.
In a recent work, \citet{atrias} uses simulations to build a lower-dimensional space which is later used to learn gait parameters on the real hardware.
Although this approach was able to achieve good results for a 9-dimensional controller with only 20 iterations, 
it requires---due to its informed kernel structure---a large amount of precomputed simulator data.

%% file: background.tex
\section{Preliminaries}

\subsection{Gaussian Process Regression}
\label{section:GP}
Given a training set $\mathcal{D}=\{(\mathbf{x}_i,y_i)|i=1,\ldots,n)\}$ of $n$ observations, a Gaussian Process attempts to infer the relationship between the inputs and the targets given some \textit{prior} knowledge.
The observations are assumed to be corrupted by normally distributed noise $\epsilon \sim\mathcal{N}(0, \sigma^2)$, such that
\begin{equation}
\label{eq:y}
y = f(\mathbf{x}) + \epsilon
\end{equation}
\begin{equation}
\label{eq:GP}
f(\mathbf{x}) \sim\mathcal{GP}(\mu(\mathbf{x}),k(\mathbf{x}_i,\mathbf{x}_j))\, ,
\end{equation}
where $\mu(\mathbf{x})$ is the prior mean, which can be uniform, and $k(\mathbf{x}_i,\mathbf{x}_j)$ is the kernel also called the covariance function.
Using a kernel allows us to transform the input space into a higher-dimensional feature space such that a non-linear map from the input vector $\mathbf{x}$ to the function value $f(\mathbf{x})$ can be inferred. 
The kernel models the uncertainty of the mean estimate and encodes how similar $f(\mathbf{x})$ is expected to be for two vectors $\mathbf{x}_i$ and $\mathbf{x}_j$.
A high value of $k(\mathbf{x}_i,\mathbf{x}_j)$ would mean that the posterior value of $f(\mathbf{x}_j)$ is significantly influenced by the value of $f(\mathbf{x}_i)$.
Note that using kernels we do not need to know the shape of the corresponding feature space, because only the inner products in the input space are required.
 
\subsection{Bayesian Optimization}
\label{section:bayesian_reg}
Bayesian optimization is a gradient-free sample-efficient framework that optimizes a cost function $f(\mathbf{x})$ using statistical models.
Its goal is to find a global optimum of a cost function which is typically expensive to evaluate.
In our case, it would imply to wear-off the hardware of the robot by performing walking experiments.
This optimum is found by minimizing a posterior mean function.
Often, the mean and covariance of $f(\mathbf{x})$ are described by a Gaussian Process.
The points to evaluate $f(\mathbf{x})$ are selected through an \textit{acquisition function}, 
which also trades off exploration and exploitation, 
\ie to select promising points were the optimum might be and to reduce the uncertainty about $f(\mathbf{x})$.

A prominent example of those acquisition functions is \textit{Entropy Search} (ES)\cite{hennig2012}.
ES is based on the expected change of entropy $\mathbb{E}[{\Delta}H(\mathbf{x})]$, 
such that the point to evaluate in the next iteration is the one that offers most information (highest entropy change).
The location of the minimum is approximated by a non-uniform grid, 
which, upon convergence, will be peaked around the actual minimum.
The acquisition function for ES is defined as:
\begin{equation}
\mathbf{x}_{t+1} = \argmax_{\mathbf{x}\in X}(\mathbb{E}[{\Delta}H(\mathbf{x})])\, .
\end{equation}
The approximations to make ES computationally tractable can be found in~\cite{hennig2012}.

\subsection{Multi-Fidelity Entropy Search}
\label{section:mfes}
\begin{figure}[t]
	\centering
	\footnotesize
	\includegraphics[width=0.85\linewidth]{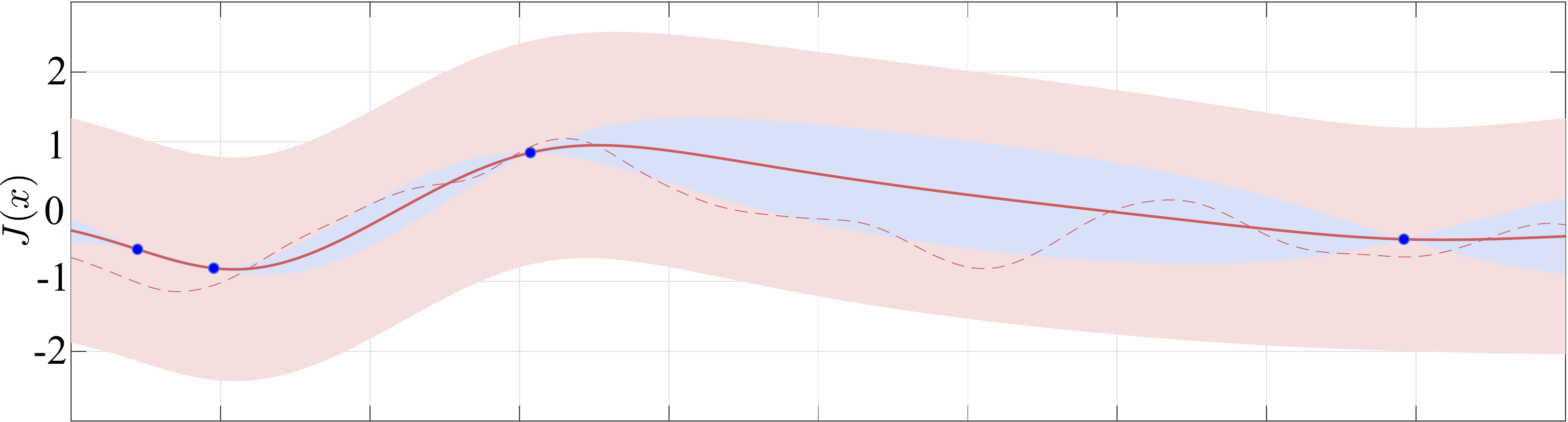}
	\includegraphics[width=0.85\linewidth]{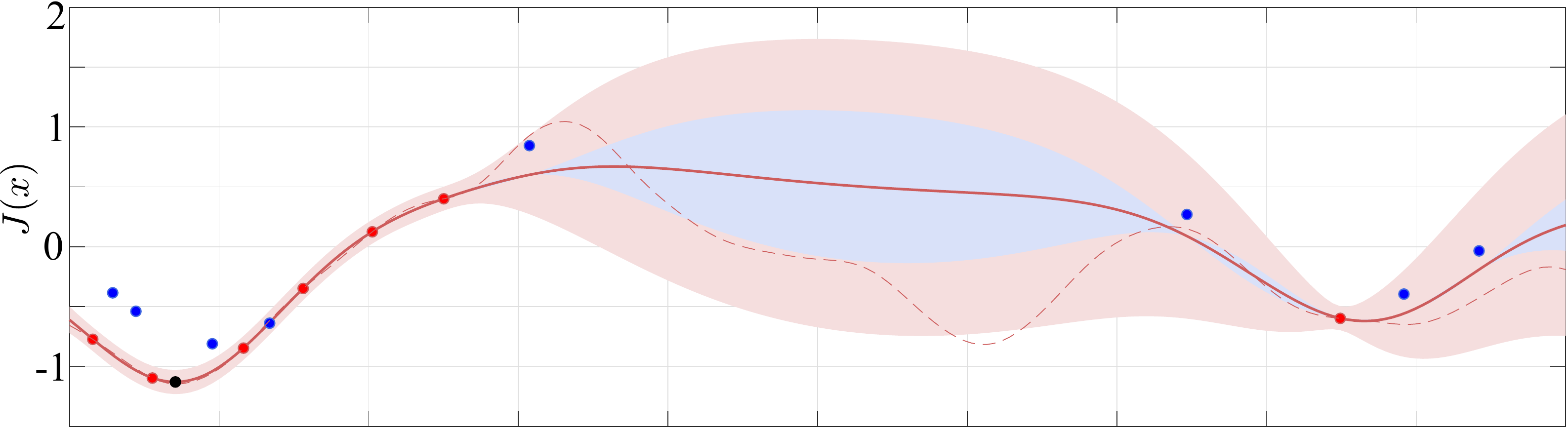}
	\caption[MultiFidelity]{Combination of simulation (blue dots) and real-robot (red dots) experiments in a synthetic example. 
		The dashed red line represents the true cost function.
		Top: simulation experiments condition the mean posterior (red line) of the cost function of the physical system $J_{real}$ and influence considerably the simulation uncertainty (blue shaded),
		but the uncertainty of the real system (red shaded) is only slightly affected.
		Bottom: this uncertainty is significantly reduced by real-robot experiments.
		Note that the difference of the simulation data points and $J_{real}$ is captured by $\epsilon_{sim}$ (Eq.~\ref{eq:J})
		The influence of simulation costs to the real system uncertainty is encoded in Eq.~(\ref{eq:kernel}).}
	\label{fig:MF_GP}
	\vspace*{-2ex}
\end{figure}

\citet{marco2017} extends the ES algorithm to integrate multiple sources of information (Fig.~\ref{fig:MF_GP}). 
The resulting method is called \textit{Multi-Fidelity Entropy Search} (MF-ES) and typically trades off real experiments with simulations. 
MF-ES optimizes the cost function
\begin{equation}
J_{real}(\mathbf{x}) = J_{sim}(\mathbf{x}) + \epsilon_{sim}(\mathbf{x})
\label{eq:J}
\end{equation}
over a parameter set $\mathbf{x}\in X$.
The key idea is to model the cost on the physical system $J_{real}$ as the cost in simulation $J_{sim}(\mathbf{x})$ plus a systematic error $\epsilon_{sim}(\mathbf{x})$.  
This error $\epsilon_{sim}$ can be a complex transformation, which is learned by the Bayesian optimization.
MF-ES defines two kernel functions $k_{sim}$ and $k_\epsilon$, which model the cost on simulation and the difference to the real experiments, respectively. 
In this manner, the kernel is expressed as:
\begin{equation}
\label{eq:kernel}
k(\mathbf{a}_i,\mathbf{a}_j) = k_{sim}(\mathbf{x}_i, \mathbf{x}_j) + k_\delta(\delta_i, \delta_j)k_\epsilon(\mathbf{x}_i, \mathbf{x}_j)\,,
\end{equation}
where $\mathbf{a}=(\delta, \mathbf{x})$ is an augmented vector in which $\delta$ indicates if a real experiment was performed and $k_\delta(\delta_i, \delta_j) = \delta_i\delta_j$ is a kernel indicator that equals one if both evaluations were performed with the real robot. 
Accordingly, two real experiments are expected to covary stronger than evaluations containing simulations.

Since $k_\epsilon$ is modeled inside the GP, we do not need to address explicitly the mapping between $J_{sim}$ and $J_{real}$; we only require assumptions about the difference between simulation versus real experiments in form of a mean and a covariance function.
Additionally, $\delta$ has to be explicitly incorporated into the acquisition function of the optimization---otherwise only real experiments would be selected, because they 
deliver more information about the target function. 
This is done by introducing weight parameters $w_i$ for both information sources. 
Thus, the acquisition function is expressed as 
\begin{equation}
\mathbf{x_{t+1}} = \argmax_{\mathbf{x}\in\mathbb{R}^d, i\in\{\textrm{sim},\textrm{real}\}} \left(\frac{{\Delta}H_t(\mathbf{x})}{w_i}\right)\,.
\end{equation}

\subsection{Bipedal Walking with Feedback Mechanisms}
\label{section:nimbrogait}
In this section, we briefly describe the gait we want optimize~\cite{allgeuer2016}.
The gait is based on an open-loop gait pattern generator as presented in~\cite{behnke2006}. 
It essentially produces high-dimensional trajectories for omnidirectional walking given a desired target velocity vector.
The open-loop gait makes use of: the joint space, the inverse, and the abstract leg  representations~\cite{missura2013}.
The abstract leg space is a representation of the leg pose, consisting of the leg extension, leg and foot angles, and is, in contrast to representations in Cartesian or joint space, designed for easy use for walking.
The gait starts from a halt pose in the abstract space and incorporates several motion primitives such as leg lifting and leg swinging, also defined in the abstract space.
The resulting pose is converted into the inverse space where more motion primitives are incorporated.
Finally, the open-loop gait outputs a joint trajectory by converting the resulting inverse pose into the joint space.

In order to improve the stability and robustness of the robot, feedback mechanisms were added to the open-loop gait~\cite{allgeuer2016}.
The orientation of the robot is represented by fused angles~\cite{fused}.
Using the deviations $d_{\alpha}$ and $d_{\beta}$ of the fused roll $\alpha_{B}$ and fused pitch $\beta_{B}$ from a desired orientation, \eg from an upright torso pose, the activation value $\mathbf{u}$ of different corrective actions is calculated.
The elements of $\mathbf{u}$ can be considered as the strength of corresponding corrective actions (feedback mechanisms) which are then applied to the open-loop gait in the abstract or inverse space.
The corrective actions include: \textit{arm swinging, hip movement, COM shifting, ankle tilting, and support foot tilting}. 
In order to obtain the activation values $\mathbf{u}$, the deviations $d_{\alpha}$ and $d_{\beta}$ are passed through a feedback pipeline composed of integrators, derivatives, mean filters and smooth deadband filters to produce a PID vector $\mathbf{e}\in\mathbb{R}^6$. 
This vector is then multiplied by a gain matrix $\mathbf{K_a} \in \mathbb{R}^{m\times6}$ to generate the activation values of the corresponding $m$ corrective actions.

%% file: approach.tex
\section{Gait Parameter Learning}
\label{section:our_approach}
As explained in Section~\ref{section:nimbrogait}, the gait is composed of two main components: an open-loop central pattern generator and feedback mechanisms.
In this paper, we address the problem of optimizing of the feedback mechanisms.
Specifically, the activation gains $\mathbf{K}_a$ of the corrective actions will be optimized.

\subsection{Cost Function}
The fused angle deviations $d_\alpha\!=\!\alpha_{des}-\alpha$ (pitch) and $d_\beta\!=\!\beta_{des}-\beta$ (roll) give us an estimate about the unintended tilt of the robot induced by walking.
These measurements are, however, very noisy. 
We apply thus a mean filter and a smooth deadband to $d_\alpha$ and $d_\beta$ as proposed in~\cite{allgeuer2016}. 
In other words, we end up with the proportional part $e_{P\alpha}$ and $e_{P\beta}$ of the fused feedback vector $\mathbf{e}$.
So, we define the stability criterion of the entire gait as the integral of $e_{P\alpha}$ and $e_{P\alpha}$ along the gait duration $T$
\begin{equation}
\int_0^T{\Vert e_{P\alpha}(\mathbf{x})\Vert_1 + \Vert e_{P\beta}(\mathbf{x})\Vert_1}dt  \,.
\end{equation}

This stability criterion will be part of our cost function.
Additionally, we introduce a penalty term that smoothly regularizes the parameter $\mathbf{x}$.
The penalty term is a logistic function of the form:
\begin{equation}
\nu(\mathbf{x})=\frac{s}{1+\exp\left( -\gamma(\Vert\mathbf{x}\Vert_2-\lambda\Vert\mathbf{x}_{max}\Vert_2) \right)}\,,
\end{equation}
where $\mathbf{x}_{max}$ is the upper bound of $\mathbf{x}$, 
$\lambda$ is a factor that affects the position of the transition, 
$s$ represents the magnitude of the penalization and $\gamma\in\mathbb{R}$ controls the smoothness of the phase transition. 

Since corrective actions in the sagittal plane are only activated by the fused angle pitch $\alpha$ and in the lateral plane by the fused angle roll $\beta$, 
we propose a cost function for the parameters $\mathbf{x}_l$ that have an effect on the lateral plane and another cost function for the parameters $\mathbf{x}_s$ that affect the sagittal plane:
\begin{equation}\label{eq:finala}
J_{\alpha}(\mathbf{x}_l) = \int_0^T{\Vert e_{P\alpha}(\mathbf{x}_l)\Vert_1}dt + \nu(\mathbf{x}_l)
\end{equation}
\begin{equation}\label{eq:finalb}
J_{\beta}(\mathbf{x}_s) = \int_0^T{\Vert e_{P\beta}(\mathbf{x}_s)\Vert_1}dt + \nu(\mathbf{x}_s)\,.
\end{equation}

In order to overcome the intrinsic error of the simulator, \ie same parameters yield different results, 
we perform $N$ evaluations in simulation with the same parameters and incorporate their mean into the cost function for the simulation:

\begin{equation}
\bar{J}_{sim,i}(\mathbf{x}) = \frac{1}{N}\left(\sum_{i=1}^{N}\int_0^T{\Vert e_{Pi}(\mathbf{x})\Vert_1}dt\right)+ \nu(\mathbf{x}),   i\in\{\alpha, \beta\}.
\end{equation}
For real-robot experiments, we set $J_{real}$ as given by Eq.~\eqref{eq:finala} and Eq.~\eqref{eq:finalb}. In both (simulations and real-robot experiments), if the robot falls, a large cost is assigned to $\mathbf{x}$.

To define the kernel function of the optimization (Eq.(\ref{eq:kernel})), we used the Rational Quadratic (RQ) kernel for $k_{sim}$ and $k_\epsilon$.
The RQ kernel introduces three parameters ($\sigma_k^2$, $\alpha$ and $\boldsymbol{l}$) to be tuned, also called hyperparameters. 
The $\boldsymbol{l}$ parameter roughly determines the distance of two points to significantly influence each other, 
the scale factor $\sigma_k^2$ determines the problem-specific signal variance, 
and $\alpha$ is a relative weight of large-scale and small-scale variations. 

\subsection{Termination Criteria}
\begin{figure}[!b]
	\footnotesize
	\centering
	\includegraphics[width=\linewidth]{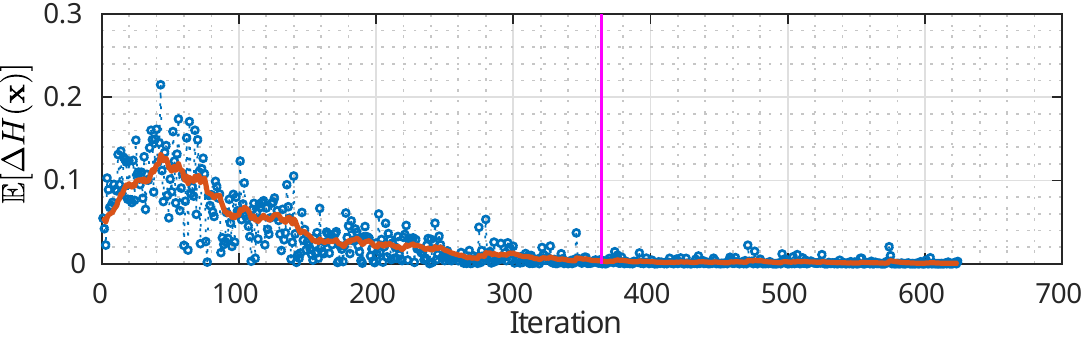}
	\caption[KLD evolution (filtered/unfiltered)]{Unfiltered (blue) and filtered with $v=0.9$ (orange) relative entropy.
		The latter is used to terminate the optimization.
		Iterations after the magenta line can be skipped because of their very low information gain.}
	\label{fig:edh_evolution}
\end{figure}

The most simple and frequently used criterion to stop global optimization algorithms is based on the number of iterations. 
Whereas this condition works fine for problems that are fast to compute, it looses applicability when iterations become more expensive.
We propose a termination criterion which is based on entropy. 
We formulate a criterion that stops the algorithm as soon as the relative entropy $\mathbb{E}[\Delta H(\mathbf{x}_t)]$ reaches a defined value. 
Moreover, to ensure that outliers do not lead to a premature stop, we apply a saturated filter to the relative entropy (Fig.~\ref{fig:edh_evolution}) defined for each iteration $t$ with a velocity factor $0<v<1$ as:
\begin{equation}
\label{eq:filter}
(1-v)\mathbb{E}[{\Delta}H(\mathbf{x}_{t-1})]+v \mathbb{E}[{\Delta}H(\mathbf{x}_t)].
\end{equation}

This criterion is applied after a minimum number of iterations since a bad prior mean can lead to a low relative entropy right after the first iteration.
The entropy termination criterion is also combined with a maximum number of iterations criterion in case the relative entropy threshold is not reached mainly because the prior assumptions might not represent reality well enough. 

%% file: experiments.tex
\section{Evaluation}
\label{section:results}

\subsection{Experiment Setup}
\label{section:setup}
\subsubsection{Robot Platform}
We test our approach on the \iguhop \cite{Allgeuer2015b}. 
The robot has in total 20 degrees of freedom: 6 for each leg, 3 for each arm and 2 for the neck.
The links of this platform are fully 3D printed.
The robot is \unit[92]{cm} tall and weights \SI{6.6}{\kilogram}.
The platform incorporates an Intel Core i7-5500U CPU running a 64-bit Ubuntu OS and a Robotis \cm microcontroller board, which electrically interfaces with its Robotis Dynamixel MX actuators. 
A 3-axis accelerometer and gyroscope sensors are also contained in the \cm, for a total of 6 axes of inertial measurement. 

\subsubsection{Software Architecture}
Due to the limited computational power of the robot, 
the optimization and simulations are performed on a desktop PC with a Core i7-4890K CPU and 8\,GB of RAM. 
The simulations are carried out in Gazebo 2.2 with a real-time factor of 1.5.
We use ODE as the dynamics engine without constraint force mixing and an error reduction parameter of 0.2.
The geometries of the joint links are approximated using convex hulls.
The joints are controlled using the \emph{ros\_control} package.
To avoid induced noise in the simulation, the simulator is reset after each performed trial.
The Bayesian optimization run in Matlab 2017b.
All the components required for the gait are implemented in C\texttt{++} and executed on the robot's computer.
The interprocess communication is implemented using the ROS middleware.

\subsubsection{Scenarios}
We propose two scenarios to evaluate our approach. 
We initially perform a 2D optimization to learn the P- and D-gain of the \textit{Ankle Tilt} corrective action.
In the second scenario, a 4D optimization to learn the P- and D- gains of the \textit{Swing Arm} and \textit{Ankle Tilt} actions is performed. 
In both scenarios, there are no other feedback mechanisms active.
To avoid artifacts coming from the transient of a robot in a static configuration, 
all experiments start with the robot walking on the spot.
The walking sequence is then defined as walking on the spot during three seconds and then walking forward with a speed of \unit[0.3]{m/s}.
We do not bias the optimizer specifying any initial values of the parameters. 
We compare the results of our algorithm against manually tuned parameters devised by experts.
These parameters were used by the winning team NimbRo TeenSize at the RoboCup 2017 competition \cite{Rodriguez}.

\subsubsection{Parameters}
We parametrize the kernel function with $\boldsymbol{l}=\left( \frac{\mathbf{x}_{max}}{8} \right)$ and $\alpha = 0.25$ to produce a reasonable trade-off between exploration and exploitation.
We use the same values of $\boldsymbol{l}$ and $\alpha$ for $k_{sim}$ and $k_\epsilon$. 
The standard deviation of $k_{sim}$ is set to $\sigma_{sim}=2.48$ and the standard deviation of $k_\epsilon$ is set to $\sigma_{\epsilon}=2.07$ for the 2D optimization, whereas $\sigma_{sim}=2.07$ and $\sigma_{\epsilon}=1.79$ are set for the 4D optimization.
The prior means are set to $\mu_{sim}=53.3502$ and $\mu_{\epsilon}=-37.1385$.
These values are chosen from initial experiments.
Additionally, the penalty function $\nu(\mathbf{x})$ is parametrized with $\lambda = 0.75$ and $s=7.5$ to punish parameters larger than $\lambda\mathbf{x}_{max}$. 
The smoothness of the phase transition is set to $\gamma=6$.
The effort of the simulation and real-robot experiments are set to $w_{sim}=10$ and $w_{sim}=50$, respectively, \ie a real-robot walking sequence is five times as expensive as a simulated one.

\subsection{Experimental Results}

\subsubsection{2D Optimization}
\begin{figure}[!b]	
	\hspace{11ex} Iteration 60 \hspace{28ex} Iteration 126 \\ 
	\includegraphics[height=0.25\textheight]{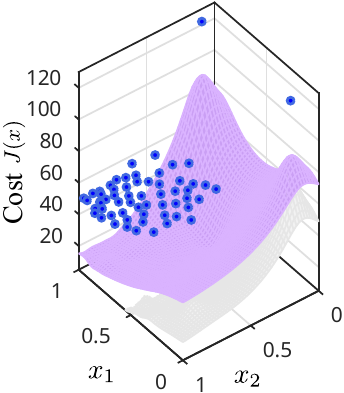}
	\includegraphics[height=0.25\textheight]{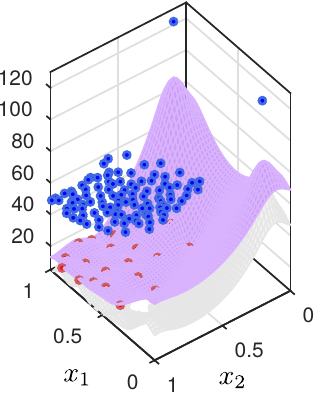}
	\includegraphics[height=0.25\textheight]{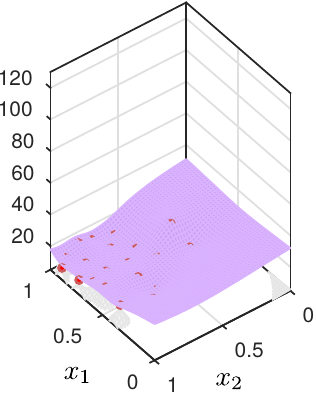}
	\caption[2D MF-ES Gaussian Process]{Optimization of the parameters for the \textit{Ankle Tilt} corrective action. 
		The purple mesh resembles the posterior mean while the gray mesh shows the covariance.
		At the first iterations, the algorithm decided to perform only simulations (blue). 
		Once enough information has been gathered in the simulation, real experiments (red) are carried out.
		To stress the contribution of the simulation, the rightmost plot shows the posterior considering only real-robot experiments.}
	\label{fig:2dGP}
	\footnotesize
	\vspace{-2ex}
\end{figure}

The algorithm performed in total 126 iterations, from which 20 walking trials were carried out with the real robot. 
This implies that approximately one real robot walking trial was required for every five simulations, 
which shows the applicability of the integration of simulation and real robot experiments for learning the activation value of the \textit{Ankle Tilt} corrective action. 
The optimized values yielded a cost of $9.3$, while the manually tuned ones resulted in a cost of $13.77$.
Thus, the optimization process found parameters that are approximately 32\% better than the manually by-expert-tuned parameters.
The resulting posterior is depicted in Fig.~\ref{fig:2dGP}.
The difference in the cost of the simulation compared to the cost with the real robot is mainly caused by the noise of the simulator, 
\eg by modeling errors of the floor impacts.

Our approach was compared with a Random Search algorithm that greedily searches for the optimal value by corrupting the current best guess with uniform noise at each iteration. 
We applied this algorithm on the same experimental setup and a maximum of 25 real-robot experiments. 
With a cost of $11.04$, the parameters of the random search outperformed the manually tuned ones.
However, our method performed 20 real experiments and yielded $15\%$ better results than random search.
Additionally, the evaluations of the random search caused one fall of the robot which never occurred with our approach. 

\subsubsection{4D Optimization}
For the 4D optimization, 301 iterations were carried out: 271 in simulation and 30 with the real robot,
\ie in average one real-robot experiment is required for every nine simulations. 
The resulting parameters were evaluated against manually tuned parameters, 
performing with the real robot 15 walking sequences for each set of parameters.
The optimized parameters yielded an average cost of 10.38 and resulted, in comparison to the manually tuned gait with a mean cost of 16.28, in an improvement of 35\%. 
In order to evaluate only the contribution of the stability in the cost, 
we subtract the penalty term of the cost; 
the resulting performance of the optimized parameters is 53\% better than the manually tuned parameters.
\begin{figure}[!b]
	\centering
		\includegraphics[height=0.35\linewidth]{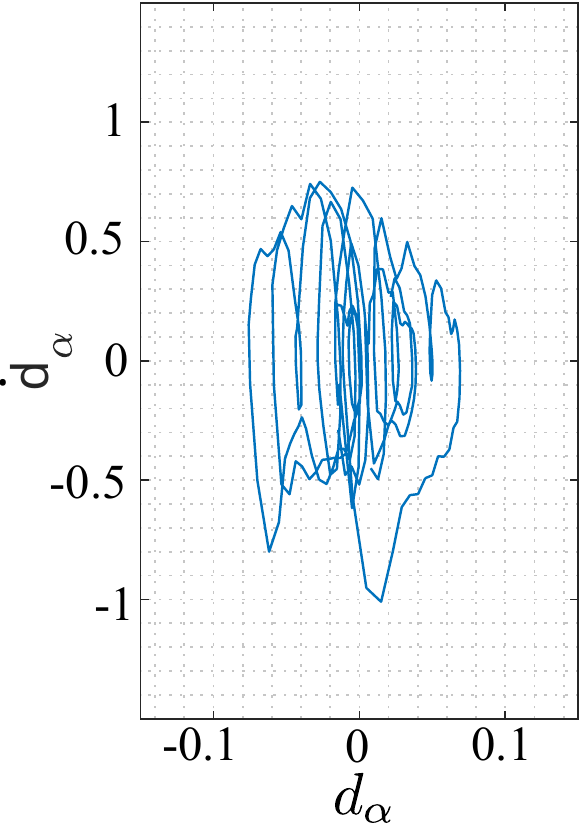}
		\includegraphics[height=0.35\linewidth]{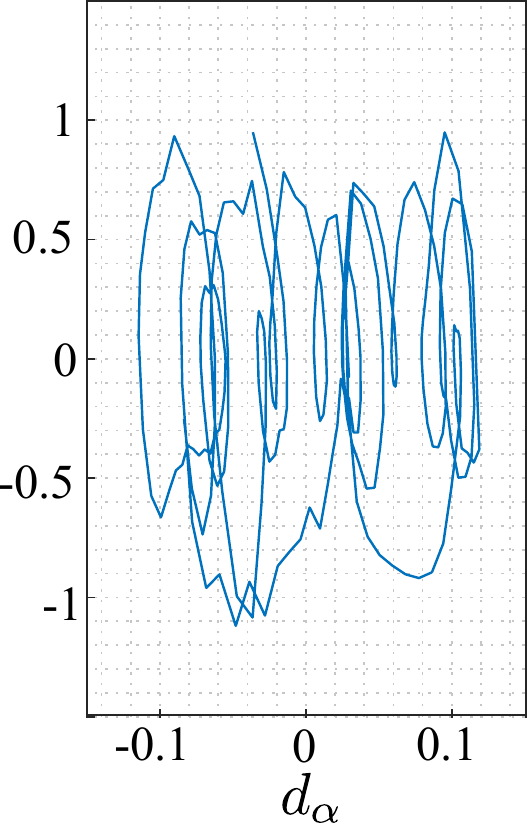}
		\includegraphics[height=0.35\linewidth]{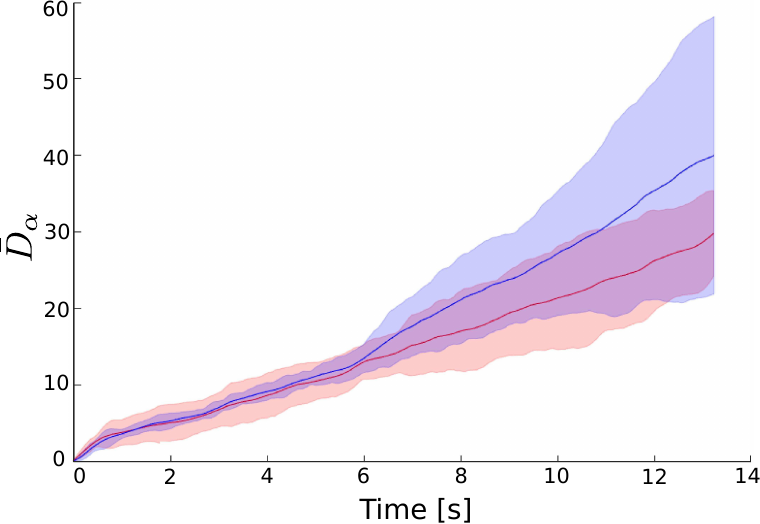}
	\caption[4D phase plots]{Phase plots of the optimized (left) and manually (middle) tuned parameters for our gait. For clarity, only one walking sequence is displayed but all the evaluated sequences show a similar behavior. Right: integral of the mean absolute fused angle deviation $\bar{D}_\alpha$ of the optimized (red) and the manually tuned (blue) parameters. The shaded regions cover the values within two standard deviations ($\pm 2\sigma$).}
	\label{fig:phase_plot}
\end{figure}

Moreover, we compared the measured fused angle deviation (not fused angle feedback) and its integral during five trials. 
As expected, during the first three seconds (walking on spot) there are no significant differences.
However, as soon as the robot starts walking forward, the deviations start to diverge and the difference between the set of parameters becomes apparent.
In general, the optimized parameters reproducibly generate deviations of lower amplitude. 
The difference becomes more apparent observing the integral of the mean absolute deviation $\bar{D}_\alpha=\int_0^T \mathbb{E}\left[\Vert d_\alpha\Vert\right] dt$ depicted in Fig.~\ref{fig:phase_plot}.

A remarkable property of our approach is the fact that the real robot did not fall a single time during the optimization process, 
because parameters that resulted in a fall in simulation were ruled out without the need for real-robot experiments.
Furthermore, the optimized gait looks qualitatively more stable and generally walks with a more upright torso compared to the manually tuned parametrization.
A video of the gait with the optimized parameters is available online\footnote{\url{http://www.ais.uni-bonn.de/videos/RoboCup_Symposium_2018}}.
The optimized gait was also tested in a very rough terrain, where the robot successfully traversed a series of debris (Fig.~\ref{fig:phase_plot}).
 
\begin{figure}[t]
  \centering
  \subfloat[]{
    \includegraphics[height=0.23\linewidth]{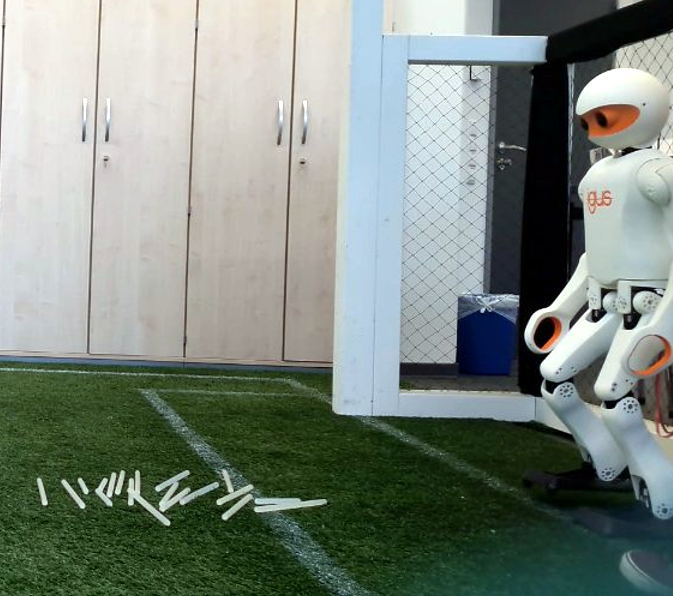}
    \includegraphics[height=0.23\linewidth]{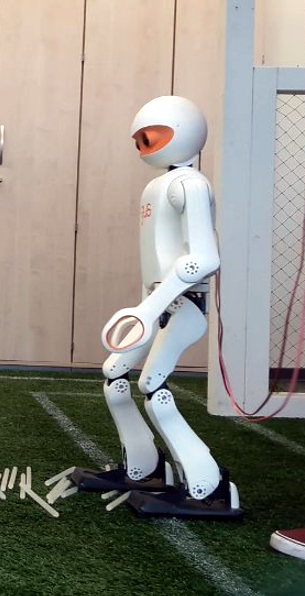}
    \includegraphics[height=0.23\linewidth]{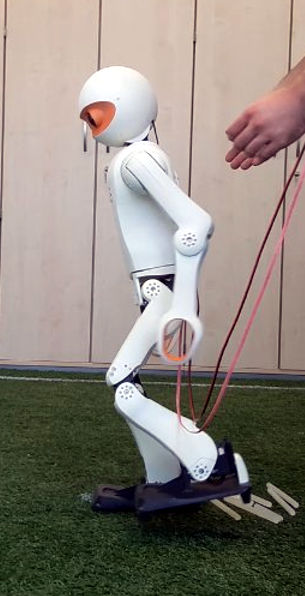}
    \includegraphics[height=0.23\linewidth]{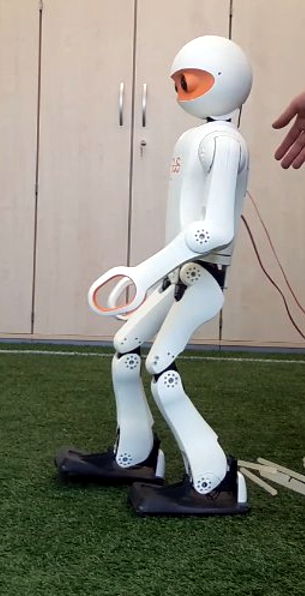}
  }
  \subfloat[]{
    \includegraphics[height=0.24\linewidth]{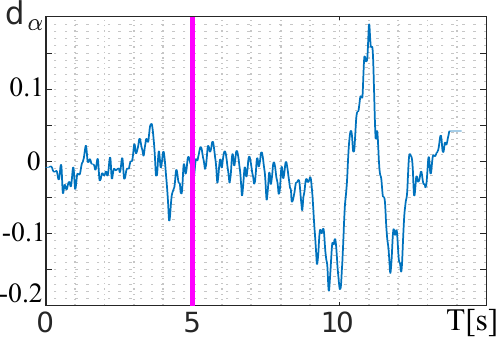}
  }
  \caption[Obstacle Experiment]{Evaluation of the optimized gait parameters on very rough surfaces (artificial grass with small debris). a) Evaluation setup (leftmost) and pictures of the robot traversing the debris. b) The corresponding fused angle deviation over time. The magenta line indicates the moment of contact with the debris.}
  \label{fig:gaitcmp_long}
\end{figure}

%% file: conclusions.tex
\section{Conclusions}
We presented an approach to trade off simulations and real-robot experiments for learning gait parameters based on a state-of-the-art Bayesian optimizer.
We showed how the gait stability was improved with the parameters found by our approach.
During the optimization process, the real robot did not fall a single time, 
which shows that the algorithm was successfully generalizing the information gathered from the simulation. 
This generalization also leads to a lower number of required physical experiments, which enables the applicability of our approach.

We observed a limitation of our method to be applied in higher dimensions. 
We hypothesize to solve this issue by using dimensionality reduction methods and by performing the optimization in a lower-dimensional space.
Additionally, in the future, we also want to learn the hyperparameters of the kernel during optimization.